\documentclass[11pt]{article}

\usepackage[utf8]{inputenc}
\usepackage[T1]{fontenc}
\usepackage{lmodern}
\usepackage[margin=1in]{geometry}
\usepackage{amsmath,amssymb,amsthm}
\usepackage{amsmath}
\usepackage{amsthm}
\usepackage{bm}
\usepackage[overload]{empheq}  

\usepackage{tabularx}
\usepackage{longtable} 
\usepackage{colortbl}

\usepackage[ruled,vlined]{algorithm2e}
\LinesNumbered

\usepackage{bm}
\usepackage{graphicx}
\usepackage{booktabs}
\usepackage{multirow}
\usepackage{array}
\usepackage{float}
\usepackage{caption}
\usepackage{subcaption}
\usepackage{xcolor}
\usepackage{siunitx}
\usepackage{enumitem}
\usepackage{CJKutf8}

\graphicspath{{Images/}}
\usepackage{cite}

\usepackage[
    colorlinks=true,
    linkcolor=black,
    citecolor=black,
    urlcolor=black,
    anchorcolor=black
]{hyperref}

\usepackage{cleveref}

\title{\textbf{Uncertainty-Aware Continual Learning for Open-World Intent Discovery Under an evolving Label Space}}

\author{
Aida Pisante\textsuperscript{*}
\and
Simone Formentin\textsuperscript{*}
}

\date{
\small
\texttt{aida.pisante@mail.polimi.it} \qquad
\texttt{simone.formentin@mail.polimi.it}
\\[0.5em]
\footnotesize
\textsuperscript{*}Dipartimento di Elettronica, Informazione e Bioingegneria,
Politecnico di Milano, Milano, Italy.
}
\begin{document}
\maketitle

\begin{abstract}
Real-world intelligent systems increasingly operate under open-world conditions, where user intents are not fixed or exhaustively known a priori and may evolve as new interaction patterns emerge. This paper proposes a unified uncertainty-aware probabilistic framework for continual new intent discovery under an evolving label space. Each utterance is encoded through an adaptive $\beta$-VAE into a latent mean, used for classification and density modelling and a posterior uncertainty estimate acting as a global reliability signal. Classifier confidence, posterior uncertainty and DP-GMM likelihood are combined through a multi-signal decision mechanism to distinguish known intents from potentially novel samples. Candidate novel instances are clustered through a density-based discovery module and only reliable clusters are promoted to new labels, enabling controlled label-space expansion. Replay and Elastic Weight Consolidation mitigate catastrophic forgetting and preserve previously acquired knowledge. The paper formalises continual intent discovery as a structured multi-phase open-world problem, introduces adaptive label-space expansion under stability--plasticity constraints and uses posterior uncertainty to regulate trusted-sample selection, pseudo-labelling, novelty admission and replay. Experiments show high novelty precision, stable adaptation across sequential phases and limited forgetting. Near-zero NMI and ARI indicate limited reconstruction of the complete fine-grained intent taxonomy, consistent with the framework's conservative promotion strategy. Qualitative analyses nevertheless reveal dense and locally coherent semantic clusters, showing that reliable novel structures can be discovered without exhaustive recovery of the underlying taxonomy.

\end{abstract}

\section{Introduction}
\label{sec}
Intelligent dialogue systems support natural-language interaction in applications such as customer service, virtual assistants and intelligent automation. New Intent Discovery (NID) allows these systems to identify and group previously unseen user requests into coherent semantic categories, progressively extending the initial intent taxonomy (Figure~1).
\begin{figure}[h]
    \centering
    \includegraphics[width=0.6\textwidth]{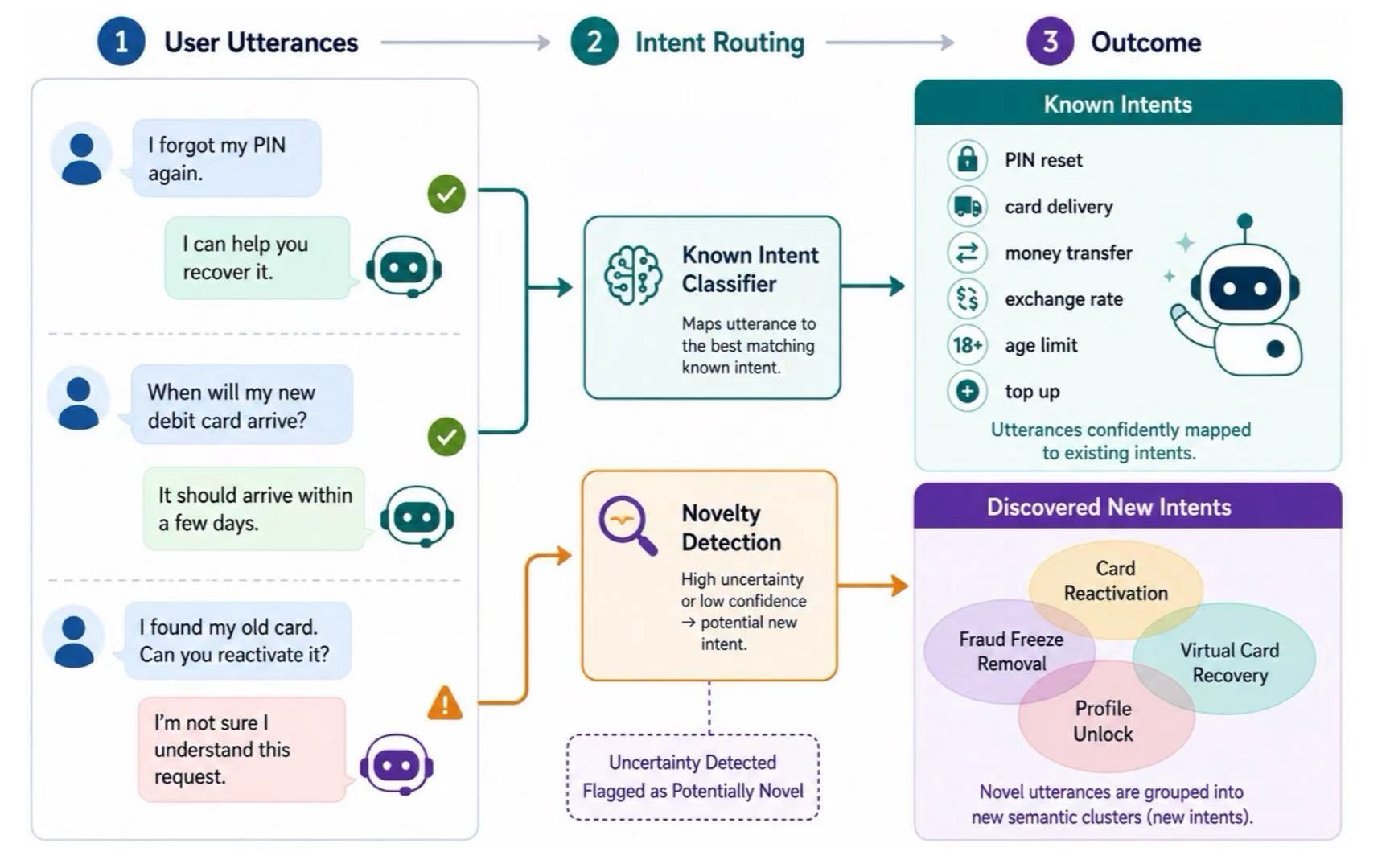}
    \caption{Intent discovery process.}
    \label{fig:intent-discovery}
\end{figure}
\\Recent research has proposed several representative frameworks. In particular \textbf{GID}\cite{mou-etal-2022-generalized} (Mou et al.(2022)) recognises labelled in-domain intents and clusters unlabelled out-of-domain utterances focusing on a single discovery stage. \textbf{RAP}\cite{zhang-etal-2024-new} (Zhang et al.(2024)) improves semi-supervised NID through prototype-based compactness and inter-class 
separation. \textbf{CGID}\cite{song-etal-2023-continual} (Song et al.(2023)), incrementally expands the classifier through prototype-guided pseudo-labelling, replay memory and feature distillation. \textbf{CDI}\cite{rawat-etal-2023-controllable} (Rawat et al.(2023)) combines MPNet representations, contrastive learning, pseudo-label refinement, Learning without Forgetting and human feedback to support controllable discovery.
Despite these advances, existing methods either assume a fixed class set, rely on deterministic representations, or lack a principled consolidation mechanism across sequential phases. This paper addresses this gap by reconceptualising continual NID as an adaptive, unified \textit{discovery--consolidation cycle}. Specifically, this framework governs the evolution of the intent label space under open-world conditions through three coordinated operations: (i) posterior uncertainty regulates information flow across phases by determining the reliability of latent representations; (ii) multi-signal density-based discovery evaluates whether emerging semantic structures satisfy the reliability, support and coherence requirements for promotion to new intent labels; (iii) continual consolidation preserves prior knowledge as the classifier and latent space adapt to the expanded taxonomy. Within this formulation, the $\beta$-VAE, DP-GMMs, experience replay and EWC act as integrated mechanisms supporting a single system-level objective: the controlled, reliable and persistent evolution of the class space over time.

\section{Uncertainty-Aware Continual Discovery Framework}
\subsection{Multi-Phase Continual Learning Pipeline}
The framework is organised as a multi-phase continual learning pipeline designed to progressively expand the intent space while preserving previously acquired knowledge. The continual setting is simulated by partitioning the full intent space into $50$ initially known intents and $50$ novel intents, which are processed across three sequential phases.

In \textbf{Phase~1}, the model is trained in a fully supervised regime on a labelled dataset: $\mathcal{D}_0
=
\left\{
\left(
\mathbf{x}_i,
y_i
\right)
\right\}_{i=1}^{N_0},$ where each utterance is encoded into a probabilistic latent representation through the $\beta$-VAE. A classification head is jointly trained to assign samples to the initial known-intent set $\mathcal{K}_0$. At the end of this phase, representative samples are stored in the replay buffer to support subsequent knowledge retention.

In \textbf{Phase~2}, the model enters the discovery stage and processes an unlabelled stream $\mathcal{D}_t^{\mathrm{unlabelled}}
=
\left\{
\mathbf{x}_i
\right\}_{i=1}^{N_t},$
which contains a mixture of previously known and unseen intents. Each sample is encoded by the $\beta$-VAE and evaluated through the multi-signal novelty-detection layer. Samples satisfying the joint novelty criterion are passed to a dedicated discovery DP-GMM, which groups them into candidate clusters; only clusters satisfying the support, density, coherence and uncertainty-based promotion criteria are retained for subsequent label-space expansion.

In \textbf{Phase~3}, the framework performs continual consolidation and label-space expansion. The promoted novel clusters are incorporated as new intent labels and the model is retrained on the current data together with pseudo-labelled novel samples and replayed past examples. The label space is consequently updated as $\mathcal{K}_{t+1}
=
\mathcal{K}_t
\cup
\mathcal{K}_{\mathrm{new}}.$ Replay and EWC preserve previously learned representations while the classifier adapts to the expanded intent space.

To more closely approximate real-world deployment conditions, the architecture is further extended to a \textbf{five-stage configuration}. In this setting, Phase~1 is executed once to establish the initial knowledge base, whereas Phase~2 ans Phase~3 are repeated for two successive novel-intent batches, $\left|\mathcal{N}_1\right|=\left|\mathcal{N}_2\right|=50$. The label space therefore expands progressively after each batch, making the stability--plasticity trade-off more challenging across successive updates (see Algorithm 1).

\begin{algorithm}[!t] \caption{Uncertainty-Aware Continual NID Framework} \label{alg:continual_nid_framework} \small \KwIn{ Initial labelled data $\mathcal{D}_0$; unlabelled batches $\{\mathcal{N}_b\}_{b=1}^{B}$; replay buffer $\mathcal{M}$ } \KwOut{Updated model and expanded label space $\mathcal{K}$} Initialize the $\beta$-VAE, classifier, DP-GMMs and label set $\mathcal{K}\leftarrow\mathcal{K}_0$\; \textbf{Phase 1: Known Intent Learning}\; Train the $\beta$-VAE and classifier on $\mathcal{D}_0$\; Encode samples into latent means $\mu_i$ and uncertainties $\sigma_i$\; Select trusted samples using $\sigma_i \leq \tau_{\sigma}$\; Fit the known-intent DP-GMM on trusted latent representations\; Initialize replay buffer $\mathcal{M}$ and EWC parameters\; \For{each unlabelled batch $\mathcal{N}_b$}{ \textbf{Phase 2: Novelty Detection and Discovery}\; Encode samples and compute classifier confidence $c_i$, uncertainty $\sigma_i$ and DP-GMM likelihood $\ell_i$\; Select novel candidates satisfying $c_i < \tau_c$, $\sigma_i > \tau_{\sigma}$ and $\ell_i < \tau_{\ell}$\; Fit the discovery DP-GMM on candidate samples\; \For{each candidate cluster $C_k$}{ \If{ $|C_k| \geq n_{\min}$ \textbf{and} $\bar{\ell}_k \geq \tau_{\ell}^{\mathrm{disc}}$ \textbf{and} $\bar{\sigma}_k \leq \tau_{\sigma}^{\mathrm{disc}}$ }{ Promote $C_k$ to a new intent label\; } } \textbf{Phase 3: Label Expansion and Consolidation}\; Expand the classifier and update $\mathcal{K}$\; Train on current and replay samples using classification, reconstruction, KL and EWC losses\; Update $\mathcal{M}$ and refit the known-intent DP-GMM\; } \end{algorithm}

\subsection{Representation Learning}

Each utterance $x_i$ is first mapped to a dense semantic embedding
$e_i \in \mathbb{R}^{d_e}$ by a pretrained sentence encoder. The embedding is then processed by a $\beta$-VAE:
\begin{equation}
q_{\phi}(z \mid e_i)
=
\mathcal{N}
\left(
\mu_i,
\operatorname{diag}(\sigma_i^2)
\right),
\label{eq:latent_posterior}
\end{equation}
where $\mu_i$ is the posterior mean used for classification and density-based modelling, whereas $\sigma_i$ quantifies the associated uncertainty for reliability-aware operations.

The $\beta$-VAE is optimized through the objective:
\begin{equation}
\mathcal{L}_{\mathrm{VAE}}
=
\mathcal{L}_{\mathrm{rec}}
+
\beta_{\mathrm{eff}}
D_{\mathrm{KL}}
\left(
q_{\phi}(z \mid e_i)
\parallel
\mathcal{N}(0,I)
\right),
\label{eq:vae_loss}
\end{equation}
where $\mathcal{L}_{\mathrm{rec}}$ encourages the decoder to reconstruct the original semantic embedding from the sampled latent variable. The KL-divergence term keeps posterior representations close to $\mathcal{N}(0,I)$, yielding a compact and stable latent space for DP-GMM modelling and continual adaptation.Its contribution is controlled by the adaptive coefficient $\beta_{\mathrm{eff}}$, which increases when the average KL divergence exceeds a target level and relaxes otherwise. 

\subsection{Uncertainty-Aware Modelling}
The posterior standard deviation is derived directly from the log-variance output of the $\beta$-VAE encoder:
\begin{equation}
\sigma_i =
\frac{1}{d_z}
\sum_{j=1}^{d_z}
\exp\left(
\frac{1}{2}
\log \sigma_{ij}^{2}
\right).
\label{eq:sample_uncertainty}
\end{equation}
It quantifies the reliability of the latent representation of each sample: low values indicate that the input is mapped to a well-defined latent region, whereas high values indicate a more ambiguous or weakly represented observation. The resulting scalar score regulates the contribution of individual samples to uncertainty-aware decisions across phases. Rather than relying on a fixed reliability threshold, the framework adopts an adaptive quantile-based criterion. The quantile level is updated after each phase according to:
\begin{equation}
q_t
=
q_{t-1}
+
\eta
\left(
r^\star - r_t
\right)
\label{eq:adaptive_quantile_update}
\end{equation}
where $q_t$ is the uncertainty quantile at phase $t$, $r^\star$ is the target trusted-sample ratio and $r_t$ is the observed trusted ratio and a sample is considered trusted when: $\sigma_i \leq \tau_\sigma$. This mechanism makes reliability a dynamic property that is recalibrated as the latent distribution evolves across phases. The resulting uncertainty signal propagates throughout the framework: it determines which samples are used to fit the known-intent GMM, filters candidates for pseudo-label assignment, contributes to novelty detection and guides replay-buffer composition by balancing reliable low-uncertainty exemplars with more ambiguous samples. By limiting the influence of unstable latent representations on density estimation, classifier updates and label-space expansion, uncertainty-aware gating reduces the risk of error propagation and supports the stability--plasticity trade-off required in open-world continual intent discovery.

\subsection{Density-Based Novelty Detection}
Novel intent discovery relies on a Dirichlet Process Gaussian Mixture Model (DP-GMM)\cite{cendra2024effective} fitted to PCA-projected $\beta$-VAE latent means. PCA\cite{zhang2026multistage} retains $95\%$ of the variance, improving numerical conditioning and stabilising covariance estimation before density modelling:
\begin{equation}
p(z)
=
\sum_{k=1}^{K}
\pi_k \,
\mathcal{N}
\left(
z \mid \mu_k, \Sigma_k
\right),
\label{eq:gmm_density}
\end{equation}
where $\pi_k$, $\mu_k$ and $\Sigma_k$ respectively denote the mixture weights, component means and covariance matrices. 
The framework employs a truncated Dirichlet Process Gaussian Mixture Model, which can activate an adaptive number of components up to a maximum value $K$, avoiding the need to specify the effective number of latent groups \textit{a priori}. This is particularly suitable for continual open-world learning\cite{shi2024continual,yang2024recent}, where new intents may progressively alter the structure of the latent space. The mixture is estimated through Expectation--Maximisation (EM) and provides soft component assignments, allowing ambiguous samples near the boundary between known and emerging regions to be represented without forcing premature hard decisions.

A non-forced assignment strategy (fallback set to ``OFF'') ensures that samples are assigned to known intents only when sufficient evidence exists; otherwise, they are forwarded as novelty candidates to the discovery module.
\\
During the initial supervised phase, a primary DP-GMM is fitted only on trusted latent representations of known intents, thereby defining a probabilistic reference model of the known-intent manifold. For each trusted sample, the model produces a log-likelihood score. Its empirical distribution is summarised through the median and Median Absolute Deviation (MAD), which define robust compatibility thresholds that are less affected by outliers than mean--variance-based statistics.

In subsequent phases, the primary DP-GMM identifies samples that are poorly explained by the known-intent density. These candidates are then analysed by a secondary DP-GMM, fitted exclusively to the candidate set, to determine whether they form sufficiently dense and coherent groups. Only clusters with adequate support, internal likelihood and reliability are promoted to new intent labels. 

\subsection{Multi-Signal Novelty Detection and Cluster Promotion}
The framework identifies potential novel samples by jointly analysing classifier confidence, $\beta$-VAE posterior uncertainty and DP-GMM likelihood under the trusted known-intent distribution. A sample is forwarded to discovery only when all three signals indicate poor compatibility with the current label space: its predicted known-intent probability is low, its latent representation is uncertain and it lies in a low-density region of the known-intent manifold. Samples with inconclusive evidence are left unassigned, avoiding forced assignments to either known or newly discovered classes.

Novelty candidates are then clustered by a dedicated DP-GMM. A cluster\cite{masuyama2023parameter} is promoted to a new intent only when it contains sufficient samples, forms a dense and coherent latent region and has low average uncertainty. This reliability-driven promotion prevents isolated outliers, fragmented groups and unstable pseudo-labels from expanding the label space. For each promoted cluster, the classifier output layer is extended by preserving the weights of existing intents and adding a new output unit, which is refined during the subsequent replay--EWC consolidation stage. The label space is therefore updated as: $\mathcal{K}_{t+1}
=
\mathcal{K}_t
\cup
\mathcal{K}_{\mathrm{new}}.$ After integration, the newly discovered intents are treated as standard known classes and contribute to replay-buffer construction, density modelling and uncertainty estimation in all subsequent phases. 

\subsection{Continual Stabilization}
To mitigate catastrophic forgetting the framework combines two complementary stabilisation mechanisms operating at different levels of the learning process. 
At the data level, the framework maintains a fixed-capacity replay buffer\cite{aghasanli2025prototype} $\mathcal{R}$ of $M=5{,}000$ samples. When the buffer is updated, it retains $95\%$ high-uncertainty (\emph{hard}) samples and $5\%$ low-uncertainty (\emph{easy}) samples. The hard subset preserves ambiguous samples near decision boundaries, where later updates are more likely to cause representational drift, whereas the easy subset provides stable prototypes of well-consolidated known-intent regions. This uncertainty-guided allocation prioritises informative boundary cases while retaining a compact reference of reliable past knowledge, supporting both plasticity and stability across continual phases.

At the parameter level, Elastic Weight Consolidation (EWC) penalises changes to parameters that were important in previous learning phases. Using a diagonal approximation of the Fisher Information Matrix, the regularisation term is defined as:

\begin{equation}
\mathcal{L}_{\mathrm{EWC}}(\boldsymbol{\theta}) =
\frac{\lambda_{\mathrm{EWC}}}{2}
\sum_{i} F_i
\left(
\theta_i - \theta_i^{\star}
\right)^2,
\label{eq:ewc_loss}
\end{equation}

where $\theta_i^{\star}$ denotes the consolidated value of parameter $i$ obtained at the end of the preceding learning phase, $F_i$ is the corresponding diagonal Fisher Information estimate and $\lambda_{\mathrm{EWC}} = 7{,}500$ controls the strength of the regularisation. Parameters associated with high Fisher values are therefore strongly constrained, whereas parameters estimated to be less important remain more flexible and can adapt to incoming data and newly discovered intents.

\section{Experimental Results}
The framework was evaluated over five independent runs using random seeds $\{42, 7, 13, 21, 123\}$ where all architectural components and hyperparameters were kept fixed.
\\
\\
Under the \textbf{three-phase protocol} (Figure 2), the framework exhibits moderate but stable classification performance (accuracy: $0.6296$--$0.6924$) alongside a conservative novelty-detection behaviour: novel precision remains consistently high ($0.7413$--$0.8791$) while novel recall spans a wider interval ($0.3898$--$0.6064$), reflecting the strict multi-signal admission criterion. Novel F1 ranges from $0.5410$ to $0.6852$. Phase-wise, both mean accuracy and Novel F1 decline monotonically reflecting structured interference rather than catastrophic forgetting, as label-space expansion progressively complicates the preservation of existing decision boundaries. This degradation is not attributable to uncertainty instability: \texttt{sigma\_quantile} and \texttt{actual\_trusted\_ratio} remain stable near $0.90$ throughout, with the adaptive quantile evolving only marginally ($0.8990 \to 0.8952$) and the $\sigma$ cutoff shifting from $1.0495$ to $0.9088$, while GMM likelihood thresholds recalibrate progressively to the expanding latent space. 
\begin{figure}[H]
    \centering
    \includegraphics[width=0.75\textwidth]{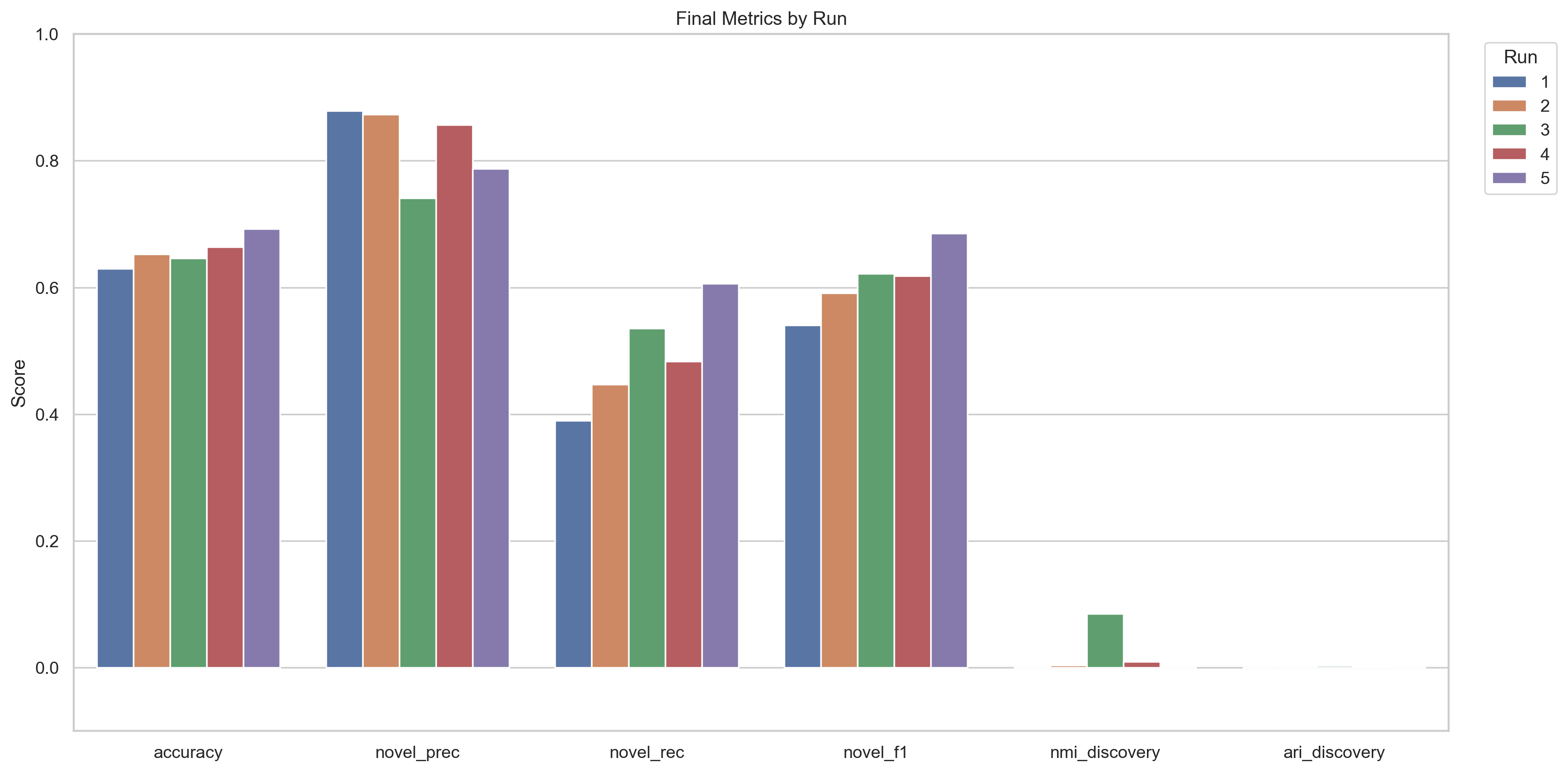}
    \caption{Three-Phases Final Metrics}
    \label{fig:intent-discovery}
\end{figure}
Global clustering metrics (NMI and ARI) remain close to zero mainly because the framework follows a conservative promotion policy, accepting only reliable, high-density novel regions. This selectivity favors precision and stability, but it also means that part of the fine-grained novel-intent structure is deliberately left unresolved rather than forcing uncertain assignments. As a consequence, the discovered partition covers only the most reliable regions of the novel space and does not fully reproduce the complete ground-truth taxonomy. Since NMI and ARI measure global agreement between the entire predicted clustering and the true class structure, they strongly penalize this partial and selective reconstruction. Therefore, values near zero reflect the cost of a reliability-oriented discovery strategy: the model sacrifices exhaustive taxonomy recovery in order to reduce uncertain promotions and limit error propagation across continual learning phases. Qualitative inspection nonetheless reveals a subset of promoted clusters with high or medium local purity, confirming that the latent space supports robust known--unknown discrimination even where global alignment metrics are low.
\\
\\
The \textbf{five-phase extension} (Figure 3) introduces a moderate accuracy cost while improving novelty-detection consistency: novel precision rises in its lower bound with reduced cross-run variability and Novel F1 occupies a higher and tighter range ($0.6378$-$0.6860$). 
\begin{figure}[h]
    \centering
    \includegraphics[width=0.75\textwidth]{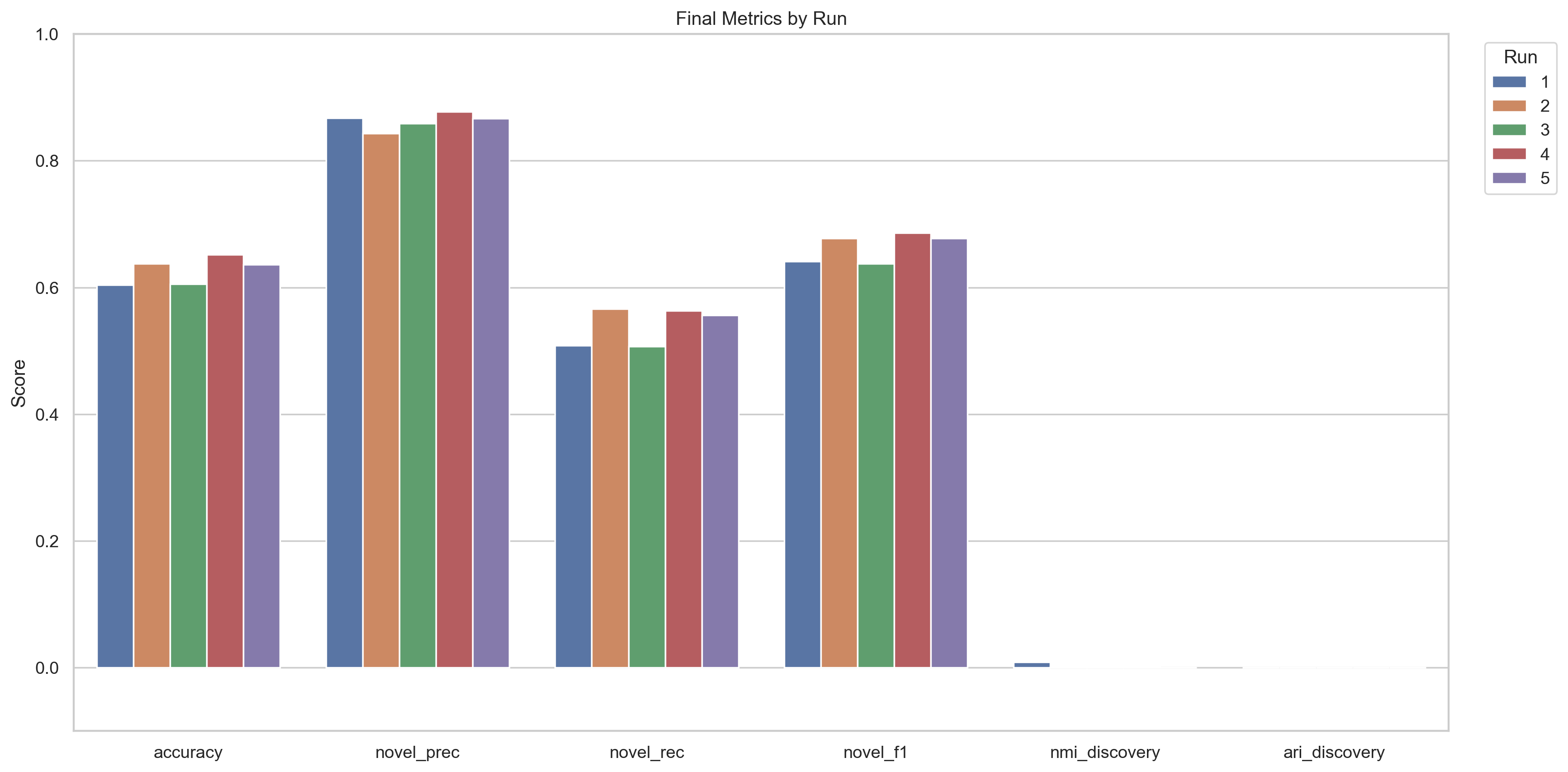}
    \caption{Five-Phases Final Metrics}
    \label{fig:intent-discovery}
\end{figure}
The phase-wise behaviour differs qualitatively (Figure 4, Figure 5): accuracy drops sharply at the first supervised-to-open-world transition ($0.707 \to 0.640$) but stabilises through Phases~3--5 and Novel F1 plateaus near $0.595$ indicating that distributing novel-intent arrival across incremental updates allows replay and EWC to absorb the adaptation burden more gradually.
\begin{figure}[ht]
    \centering
    \includegraphics[width=0.6\textwidth]{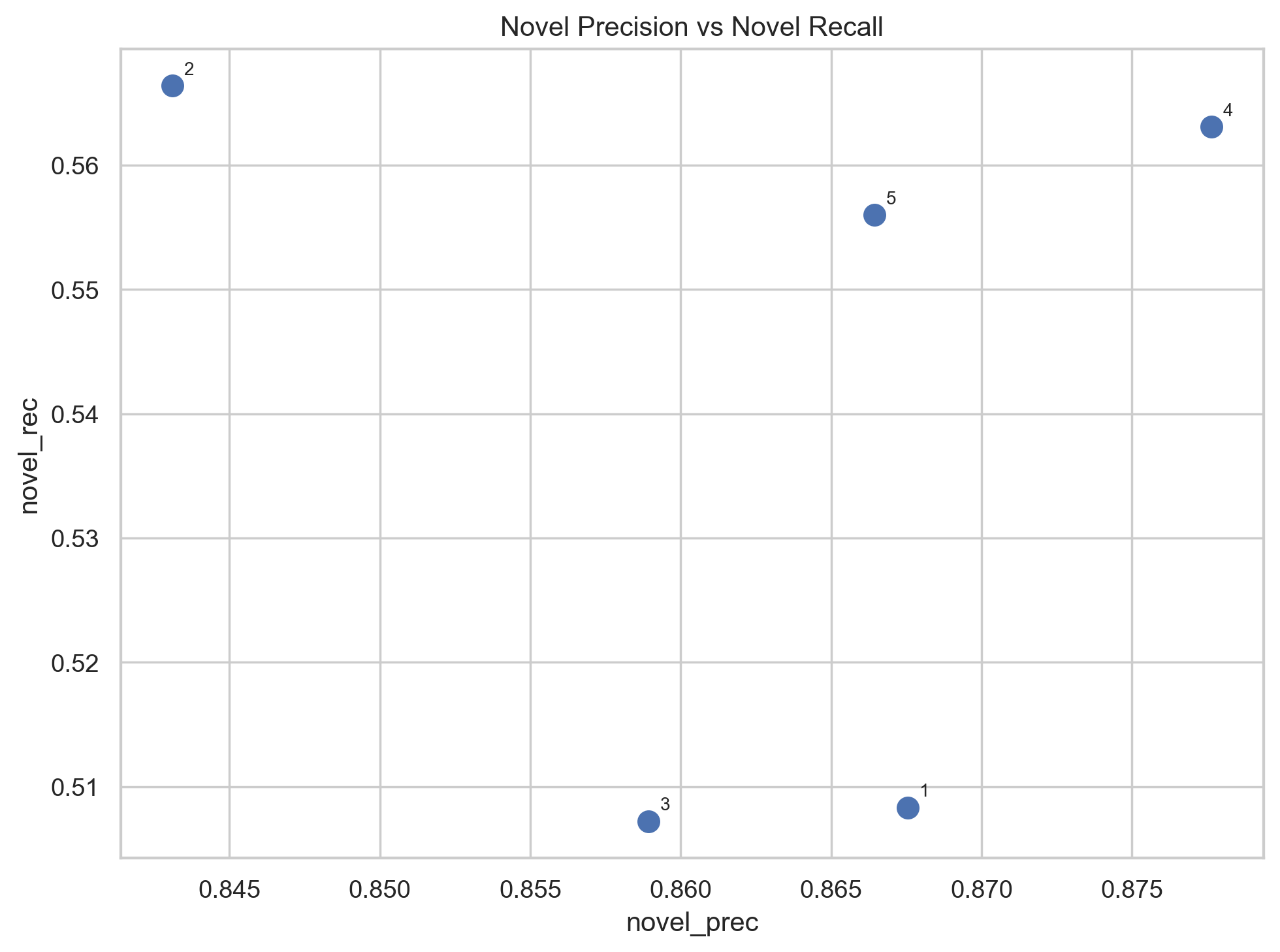}
    \caption{Novelty detection analysis}
    \label{fig:intent-discovery}
\end{figure}
Uncertainty-based filtering remains stable over all five phases: the gating mechanism is reliable under repeated label-space updates while the main adaptation burden is absorbed by the density-based decision boundaries. 
\begin{figure}[h]
    \centering
    \includegraphics[width=0.6\textwidth]{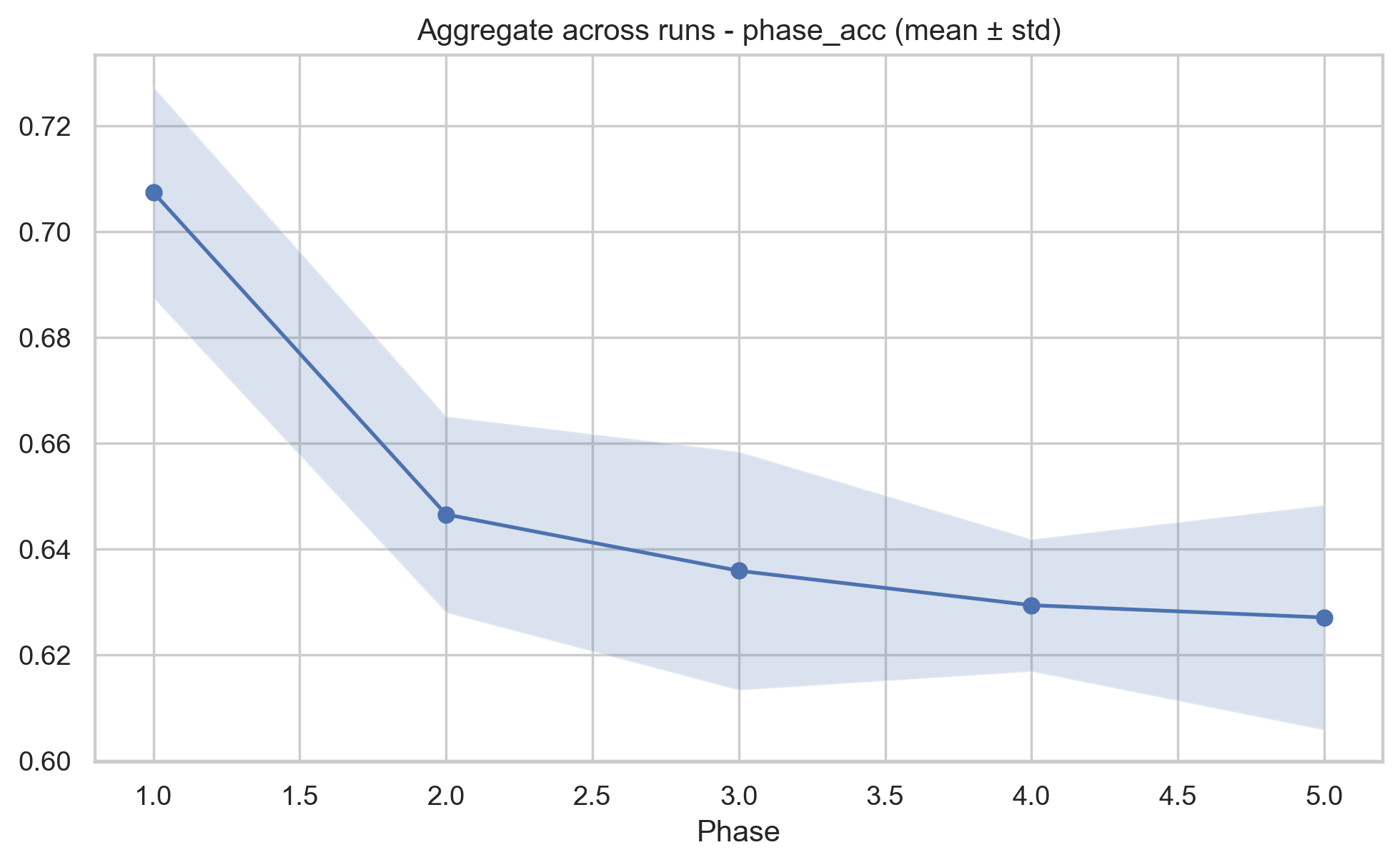}
    \caption{Continual learning across phases}
    \label{fig:intent-discovery}
\end{figure}
As in the three-phase setting, NMI and ARI remain essentially zero, with promoted cluster purity ranging from $0.91$--$0.97$ for the most coherent structures to $0.19$--$0.60$ for more mixed ones.

Table 1 summarizes the comparison of final results between the three-phase and five-phase frameworks.
\begin{table}[H]
\centering
\footnotesize
\caption{Comparison of final metrics.}
\label{tab:results}
\begin{tabular}{lcc}
\toprule
\textbf{Metric} & \textbf{Three-Phase} & \textbf{Five-Phase} \\
\midrule
Global Accuracy & 0.6296--0.6924 & 0.6043--0.6521 \\
Novel Precision & 0.7413--0.8791 & 0.8431--0.8776 \\
Novel Recall    & 0.3898--0.6064 & 0.5072--0.5664 \\
Novel F1        & 0.5410--0.6852 & 0.6378--0.6860 \\
NMI             & 0.0000--0.0847 & 0.0000--0.0083 \\
ARI             & 0.0000--0.0027 & $\approx 0.0000$ \\
\bottomrule
\end{tabular}
\end{table}

Two \textbf{ablation studies} validate the core design choices. \textbf{Replacing the DP-GMM with K-Means} ($K=16$) yields a comparable Novel F1 ($0.6233$ vs.\ $0.6860$) with a $19.7\%$ runtime reduction, but promotes fewer clusters (8 vs.\ 25) and lacks the probabilistic flexibility to model irregular latent density structures, while NMI and ARI remain near zero under both backends. \textbf{Enabling the GMM fallback} reassignment degrades both accuracy
($0.6521 \to 0.6280$) and Novel F1 ($0.6860 \to 0.5592$), with reductions in both precision ($-0.0481$) and recall ($-0.1413$), demonstrating that forced assignment of ambiguous inputs blurs the known--novel decision boundary.
\\
\subsection{Methodological Positioning}

Table~\ref{tab:method_comparison} contrasts the proposed framework with representative baselines. 

\begin{table}[H]
\centering
\footnotesize
\caption{Methodological positioning with State of the Art}
\label{tab:method_comparison}
\footnotesize
\renewcommand{\arraystretch}{1.12}
\setlength{\tabcolsep}{2.5pt}

\begin{tabularx}{\columnwidth}{@{}l X c X@{}}
\toprule
\textbf{Method} & \textbf{Setting} & \textbf{Unc.} & \textbf{Mechanism} \\
\midrule
GID       & Static IND/OOD       & $\times$      & -- \\
CGID      & Continual OW-NID     & $\times$      & Replay + distillation \\
CDI       & Stage-wise NID       & $\times$      & LwF distillation \\
RAP       & Static semi-sup.\ NID& $\times$      & -- \\
\textbf{Proposed} & Multi-phase Continual NID & $\checkmark$ & Replay + EWC \\
\bottomrule
\end{tabularx}
\end{table}

The framework formulates continual NID as an adaptive uncertainty-aware system that governs label-space evolution across successive updates. Posterior uncertainty, density-based discovery and replay--EWC consolidation jointly regulate how the system identifies, integrates and retains intents over time.

\subsection{Main Contributions}
The main contributions of this paper are threefold and jointly advance the formulation of New Intent Discovery (NID) in continual open-world settings. 

\textit{First}, the work formalizes \textit{multi-phase continual NID} under an evolving label space, addressing a research setting that remains largely unexplored in intent discovery and only partially investigated in related areas such as Generalized Category Discovery. Unlike conventional approaches that consider intent discovery as a static or single-stage problem, the proposed formulation assumes that previously unseen intents may emerge repeatedly over time and must be detected, discovered, integrated and subsequently retained across sequential learning phases. 

\textit{Second}, the paper introduces a unified mechanism for progressive label-space expansion under stability--plasticity constraints. Unlike incremental learning approaches that assume a fixed label space, open-set methods that detect unknown samples without converting them into stable knowledge and clustering-based approaches that identify new structures without integrating them into the model, the proposed framework explicitly validates discovered clusters and promotes reliable ones to new intent labels. This expansion process is tightly coupled with replay and Elastic Weight Consolidation (EWC), allowing the system to evolve its label space over time while preserving previously acquired knowledge. As a result, discovery becomes a continual process of structured knowledge expansion, enabling the integration of emerging semantic categories while mitigating catastrophic forgetting and maintaining stable learning in open-world settings.

\textit{Third}, the paper establishes posterior uncertainty as a first-class operational control signal throughout the continual discovery pipeline. Rather than using uncertainty solely as an auxiliary diagnostic measure, the proposed framework exploits the posterior uncertainty produced by the adaptive $\beta$-VAE to regulate trusted-sample selection, pseudo-labelling, novelty admission and replay selection across phases.

Together, these contributions define a unified perspective on continual intent discovery in which intelligent systems are designed not only to recognize known categories, but also to repeatedly identify, integrate and retain emerging intents while maintaining reliability in non-stationary open-world environments.

\section{Conclusion}
This work proposes an uncertainty-aware framework for continual new intent discovery, treating the task as an adaptive system for the controlled evolution of the label space. Posterior uncertainty estimation, density-based novelty admission and continual consolidation mechanisms are jointly designed to support the reliable identification and integration of emerging concepts across sequential learning phases. Experiments confirm stable, high-precision novelty detection, clustering metrics remain conservative by design, as only statistically reliable clusters are promoted, though qualitative analysis reveals locally coherent structures. Future work should address threshold sensitivity, limited separation of closely related intents and more adaptive consolidation strategies.

\bibliographystyle{IEEEtran}
\bibliography{bibliography}

\end{document}